\documentclass[a4paper,conference]{IEEEtran}

\usepackage{amsmath}
\usepackage{amsfonts}

\ifCLASSINFOpdf
  \usepackage[pdftex]{graphicx}
  \graphicspath{{../pdf/}{../jpeg/}}
  \DeclareGraphicsExtensions{.pdf,.jpeg,.png}
\else
\fi
\newcommand{\diff}[2]{\dfrac{\partial #1}{\partial #2}}

\begin{document}

\title{Conditional Information Gain Networks}

\author{\IEEEauthorblockN{Ufuk Can Bicici}
\IEEEauthorblockA{
Bogazici University\\
Idea Technology Solutions
\\Email: can.bicici@boun.edu.tr\\
}
\and
\IEEEauthorblockN{Cem Keskin}
\IEEEauthorblockA{
PerceptiveIO, Inc.\\
Email: cem.kskn@gmail.com}
\and
\IEEEauthorblockN{Lale Akarun}
\IEEEauthorblockA{Computer Engineering Department\\
Bogazici University\\
Email: akarun@boun.edu.tr}}

\maketitle

\begin{abstract}
Deep neural network models owe their representational power to the high number of learnable parameters. It is often infeasible to run these largely parametrized deep models in limited resource environments, like mobile phones. Network models employing conditional computing are able to reduce computational requirements while achieving high representational power, with their ability to model hierarchies.  We propose Conditional Information Gain Networks, which allow the feed forward deep neural networks to execute conditionally, skipping parts of the model based on the sample and the decision mechanisms inserted in the architecture. These decision mechanisms are trained using cost functions based on differentiable Information Gain, inspired by the training procedures of decision trees. These information gain based decision mechanisms are differentiable and can be trained end-to-end using a unified framework with a general cost function, covering both classification and decision losses. We test the effectiveness of the proposed method on MNIST and recently introduced Fashion MNIST datasets and show that our information gain based conditional execution approach can achieve better or comparable classification results using significantly fewer parameters, compared to standard convolutional neural network baselines.
\end{abstract}


%
\IEEEpeerreviewmaketitle

\section{Introduction}
Deep neural networks have achieved considerable success in machine learning tasks such as image classification. AlexNet \cite{Alexnet},  is the first example of a convolutional neural network (CNN) that achieved state-of-the-art performance on the ImageNet dataset. Following this seminal work, successful CNN models such as \cite{VggNet} and \cite{He_2016_CVPR} have implemented novel mechanisms to allow the training of deeper models. While such sophisticated networks increase the classification performance on various datasets, memory and computation loads render them infeasible for platforms where  resources are scarce, such as mobile phones and embedded systems.

\begin{figure}[!t]
\centering
\includegraphics[width=3.5in]{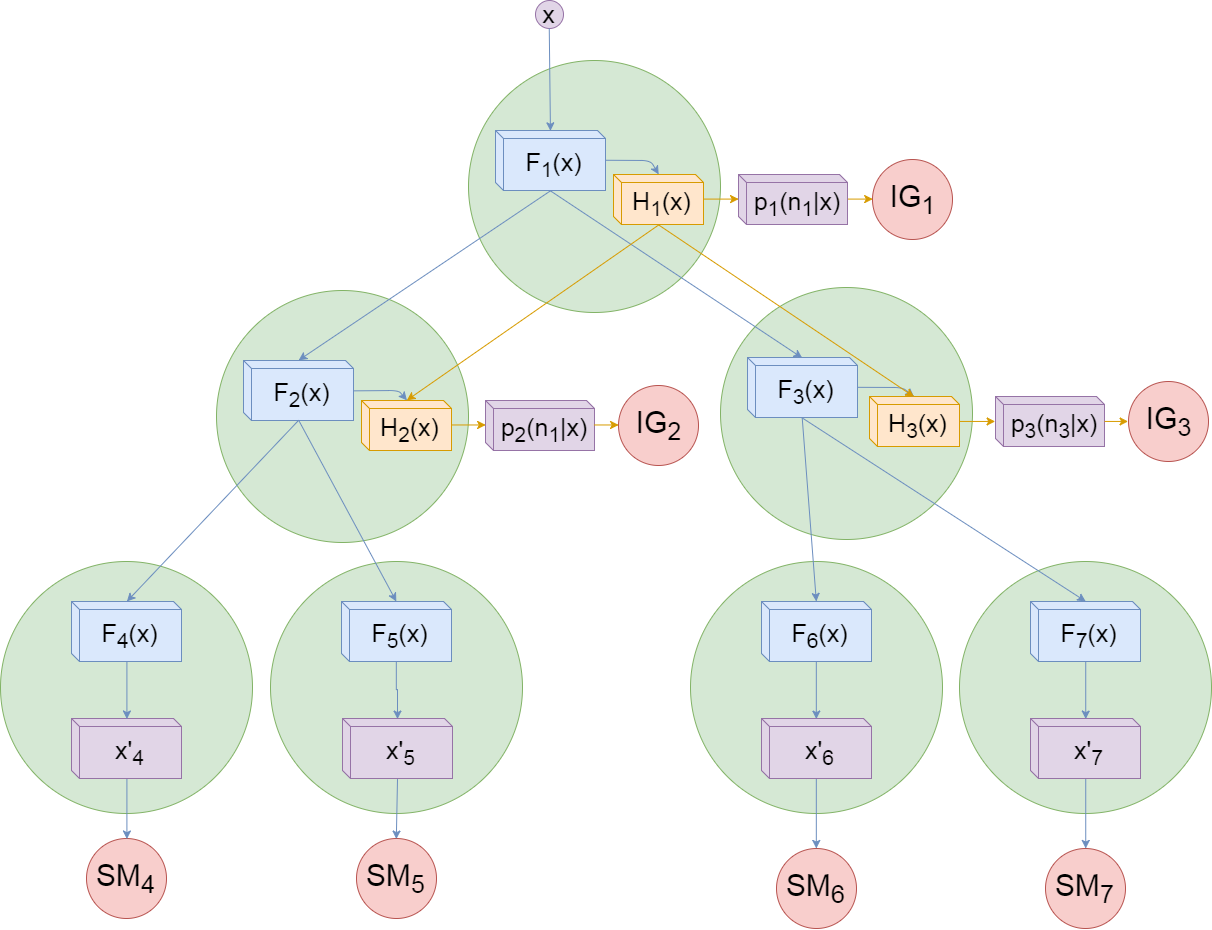}
\caption{A binary Conditional Information Gain Network with depth of two. In the split nodes, special routing networks $H_1, H_2$ and $H_3$ are trained by maximizing local information gain at these nodes and learn optimal splits, such that lower parts receive purer data. The rest of the network (consisting of usual CNN transformations, $F_i$, $1 \leq i \leq 7$) is trained with usual multinomial logistic regression loss.}
\label{fig:CIGN}
\end{figure}

Recently, conditional computing approaches have been proposed to decrease  memory and runtime requirements and to improve representational power. Conditional computing aims to disable a subset of the network, based on a given sample, both for forward calculations and/or updates during  training \cite{bengio2013deep}. It is hypothesized that the subset which remains activated is capable of correctly classifying the sample in question. In accordance with that idea, we present Conditional Information Gain Networks (CIGN), which are tree structured feed forward architectures, enabling the selection of network subsets conditioned on a given sample. Each non-leaf tree node contains two types of transformations ($F$ and $H$), driving the classification ($F$) and information gain ($H$) objectives. The information gain objective in each split node routes data to a certain branch; resulting in lower entropy compared to the whole since nodes deeper in the tree receive a subset of the data. According to our hypothesis, this will create expert sub-networks which are fine tuned to discriminate between samples belonging to a specific subset of classes. Since this supervised partitioning of the data is achieved by the router networks driven by information gain, these expert networks will have a simplified task of classification. This, in turn, allows one to reduce the number of parameters deeper in the tree while achieving better or comparable accuracy as a non-conditional baseline network with more parameters. The routing mechanism automatically achieves the goal of conditional computing;  allowing sparse update and execution of the feed forward network.

The contributions of this paper can be listed as follows:
\begin{itemize}
\item  We propose a conditional deep neural network, which allows conditional computing via sparse evaluation and updates, driven by differentiable information gain.
\item  The supervised partitioning of data into class groups generate expert sub-networks in the deeper layers of the network, which allows  decimation of the parameter count, while preserving the accuracy.
\item  We introduce counter measures to protect the tree structured network from errors made by the routers and to provide data load balance in deeper levels.
\end{itemize}

\section{Related Work}
\subsection{Conditional Computing}
Conditional computation in deep neural networks aims to reduce the burden of computation costs during evaluation and updates, via sparse activation of network parts. One of the earliest works on this line is of Bengio et al. \cite{BengioStochastic}, which proposes stochastic neurons that can be turned on/off, activating only subsets of the network. A similar approach has been followed by Murdock et al. \cite{Blockout}: In Blockout, a generalization of Dropout, chunks of weight matrices are set to zero stochastically. Bengio et al. similarly apply binary masks sampled from Bernoulli distributions based on the previous activation layer, whose parameters are learned via policy gradients using the REINFORCE algorithm \cite{Williams92REINFORCE}. Wu et al. propose Blockdrop, in which a helper network learns to selectively drop a subset of the network layers  in the ResNet architecture, using reinforcement learning \cite{wu2017blockdrop}.

While the methods above implicitly sample network structures, another approach is to impose an explicit hierarchical structure with routing mechanisms conditioned on the incoming sample. Ioannou et  al. \cite{ioannou2016decision} use small routing neural networks, which assign weights to each path, later to be merged by weighted averaging. During inference; only the paths assigned the top-k weights by the routers are selected. Xiong et al. propose a binary tree structured CNN for modality aware face recognition \cite{xiong2015conditional}. The binary tree introduces proxy split losses, which are attached to the outputs of convolutional layers in the split nodes. That loss tries to maximize the distances of the data means from a splitting hyperplane.  Liu et al. use Q-Learning for the routers to selectively drop parts of the hierarchy in a DAG network, based on a given sample \cite{Liu}. Fernando et al. build a very large DAG of smaller networks, whose connections are learned with reinforcement learning or genetic algorithms \cite{PathNet}. Denoyer et al. build a tree structured network, where specific paths from root to leaves are selected by small router networks \cite{denoyer2014deep}. The whole system is trained by using the REINFORCE algorithm. Teerapittayanon et al. introduce Branchynet, in which early exit points are inserted into a CNN architecture \cite{branchynet}.

\subsection{Mixture of Experts}
CIGN can also be viewed as a hierarchical mixture of experts (HME) model \cite{jordan1994hierarchical}. Eigen et al. extend the HME approach to a stacked model, consisting of multiple layers of gater and expert networks \cite{Eigen}. Shazeer et al. propose a sparsely gated mixture of experts model \cite{Shazeer}, using top k experts during inference and training. The HD-CNN model proposed by Yan et al. \cite{yan2015hd} explicitly learns coarse class categories via spectral clustering. A tree structured network is then built using the clustering results and fine-tuned. The approach of Ahmed et al. is also similar \cite{ahmed2016network}; they train a generalist network first, partitioning the data into coarse categories and then, the expert networks for each category are concatenated, without conditional execution.

\subsection{Neural Network - Decision Tree Hybrids}
CIGN may also be compared to neural network-decision tree hybrids.  \cite{Murthy_2016_CVPR} builds a Deep Decision Network; first classifying the data with a root network, and then using spectral clustering on the misclassified samples. New networks for each class cluster are added iteratively and trained by holding the previous networks fixed. \cite{MHeads} use parameter sharing, multiple exit TreeNets with ensemble aware losses. Recently introduced Deep Neural Decision Forests (DNDF) replace the softmax layers of CNNs with decision trees, in which the soft routing mechanisms in each split node are guided by the activations of the CNN's last fully connected layer \cite{kontschieder2015deep}. The categorical class distributions are held in the leaf nodes of these decision trees. The parameters of the guiding CNN and the class distributions in the leaf nodes are trained in an iterative fashion. This idea of unifying decision trees and CNNs has found different usages beyond image classification: Roy et al. use the DNDF framework for monocular depth estimation, reformulating it as a regression problem \cite{roy2016monocular}. Biau et al. interpret the weight connections of a neural network as a decision tree \cite{biau2016neural}. Baek et al. \cite{Baek} interpret the response maps and convolution filters in the convolutional layers of a CNN as nodes and edges of a decision jungle \cite{decisionjungle}. They define layer-wise entropy losses on probability distributions over response map-wise activation strengths of each class. Wang et al. train a tree structured neural network  \cite{wang2017using}. 

\section{Methodology}
\subsection{Architecture}
We introduce CIGN, a tree structured CNN model, consisting of a set of split nodes $N_s$ and a set of leaf nodes $N_l$. In each split node, a sample goes through two transformations. The first, ($F$) produces the usual learned representations, which are used for the multinomial logistic classifiers, commonly found in CNNs. These may contain conventional CNN layers, like convolution, pooling operations, nonlinearities, Inception modules \cite{Inception}, residual connections \cite{He_2016_CVPR}, Dropout \cite{srivastava2014dropout} and Batch Normalization \cite{BatchNorm}. The second,  ($H$) are connected to the information gain outputs and can be thought of as routers. They can also contain  common CNN layers, but are designed to be less complex. The learnable parameters in the routers ($H$) are driven by the gradients produced by the information gain outputs. By design, routers can be fed from the intermediate outputs of the $F$ transformations but they can also be independent from them, such that all the $H$ transformations constitute a separate, parallel network. Each router network defines a local probability distribution on the possible branches, conditioned on the current sample. The sample is routed down the branch with the highest probability. The parameters of the classification transformations ($F$) are learned with the gradients flowing from the classification objectives in the leaf nodes, as well as with the gradients of the information gain objectives, if the routers use $F$ outputs. The leaf nodes contain only classification transformations, which produce the logits to be used with a multinomial logistic classifier. Figure \ref{fig:CIGN} shows a binary CIGN in which routers are connected to $F$ transformations. The classification objective of the CIGN is given as:
\begin{equation}
\label{C_cost}
J_{C} = -\mathbb{E}_{p(x,y)}\left[ \log \sum_{j=1}^{|N_l|} p(l_{j}|x;W_{H},W_{F})p(y|l_{j},x;W_{F})\right] \nonumber
\end{equation}
In the above, $W_{F}$ and $W_{H}$ are the parameters for the $F$ and $H$ transformations, respectively. $l_{j}$ enumerates all paths from the root node to each of the $|N_l|$ leaf nodes. $p(y|l_{j},x;W_{F})$ is the posterior probability given the sample $x$ over the path $l_{j}$. The information gain objective, driving the routing mechanism is given as: $J_{IG} = - \lambda_{IG} \sum_{i=1}^{|N_s|}IG_{i}$, where $IG_{i}$ indicates the local information gain at the $i$th split node. $\lambda_{IG}$ is the weight term for the information gain loss. The most general form of the global objective function for a CIGN then can be given as:
\begin{equation}
\label{Global_cost}
L_{CIGN} = J_{C} + J_{IG} + \lambda_{F}||W_{F}||^2 +  \lambda_{H}||W_{H}||^2
\end{equation}
where $\lambda_{F}$ and $\lambda_{H}$ are the coefficients for the weight decay regularizers. The classification objective $J_{C}$ has the usual form of a mixture of experts model. We define a functional $\psi(.)$ over the set $P$ of probability distributions $p(l)$ with $l=1; \dots,k$, $\psi: P \mapsto \mathbb{R}^k$. $\psi$ applies one hot encoding to any input probability distribution: It sets the largest entry of $p(l)$ to $1$ and set other entries to $0$. The selection probability of an expert $l_{j}$ is defined as:
\begin{equation}
\label{Expert_Selection_Prob}
p(l_{j}|x;W_{H},W_{F}) = \prod_{i \in l_{j}}\psi(p_i(n_i|x))
\end{equation}
Here, $i$ are the indices of the split nodes on the root-leaf path $l_{j}$. The variable $n_i$ represents the paths from node $i$ to its children and hence $p_i(n_i|x)$ is the probability distribution over the child nodes of $i$, given the sample $x$. $p_i$ is produced as the last layer of the router network $H_i$ within node $i$. With the one hot transformation of these local branching distributions, we obtain a "hard" mixture of experts model. Given a sample $x$, only a single path $l_j=\hat{l}$ is active, with the probability $p(l_j=\hat{l}|x)=1$ and other paths $0$. This hard mixture of experts model naturally entails the sparse evaluation of a trained CIGN model: At each split node, the sample is routed down the path with the highest branching probability. This mechanism also allows sparse updating of the classification parameters $w \in W_{F}$ which are located on the path $l_{j}=\hat{l}$. For a single $x$, it is equivalent to updating the CNN on the path $l_{j}=\hat{l}$ as a single network with the classification loss $J_{C}$ producing gradients as:
\begin{align}
\label{C_cost_derivative}
\diff{J_{C}(x)}{w}&=-\dfrac{\partial \left(\sum_{j=1}^{|N_l|} p(l_{j}|x;W_{H},W_{F})p(y|l_{j},x;W_{F}) \right) / \partial w}{ \sum_{j=1}^{|N_l|} p(l_{j}|x;W_{H},W_{F})p(y|l_{j},x;W_{F})}\nonumber\\
&= -p(y|l_{j}=\hat{l},x;W_{F})^{-1}\diff{p(y|l_{j}=\hat{l},x;W_{F})}{w}\nonumber
\end{align}
In a minibatch setting, during the forward pass, this corresponds to applying binary masks to the current minibatch such that each minibatch subset is routed to its corresponding child node. During backpropagation, the error signals of the sparse minibatch entries are accumulated in the parent and propagated recursively up to the root node. This prevents exponential computation load both for training and inference.

\subsection{Differentiable Local Information Gain}
We update the router parameters $W_{H}$ (and  classification parameters $W_{F}$, if connected to routers) via local information gain objectives at each split node. Impurity minimization based decision tree induction is the basis of the well known ID3 algorithm \cite{quinlan1986induction}. ID3 and its variants  use greedy search in the feature space to find an optimal split, trying to minimize data impurity in the current tree node, calculating information gain by counting data points on different sides of the splitting hyperplane. Montillo et al. introduced differentiable information gain by replacing the hard split with a soft one, containing a sigmoid \cite{montillo2013entanglement}. They used the differentiable version of the information gain for recursively training decision tree nodes. Inspired by their work, we introduce the differentiable information gain as the objective function for our router networks. Let $i$ be a split node, $l_{-i}$ be the split nodes on the path from the root to the node $i$. Let $p(x)$ be the original data distribution. The data distribution at node $i$ is given as:
\begin{equation}
\label{DataDistribution}
p(x|l_{-i}) = \dfrac{p(x)p(l_{-i}|x)}{\int p(x)p(l_{-i}|x)dx} =  \dfrac{p(x)\prod_{j \in l_{-i}}\psi(p_j(n_j|x))}{\int p(x)\prod_{j \in l_{-i}}\psi(p_j(n_j|x))dx} \nonumber
\end{equation}
The conditional data distribution is the reweighting of the samples reaching node $i$, which is consistent with our sparse computation approach. We then define the following joint distribution at node $i$:
\begin{equation}
\label{Joint}
p_i(x,y,n_i) = p_i(x)p(y|x)p_i(n_i|x)
\end{equation}
Here, $p_i(x):=p(x|l_{-i})$ and $p(y|x):=\mathbb{I}\left[c(x) = y\right]$ where $c(x)$ gives the ground truth class label for $x$. $p_i(n_i|x)$ is the distribution over the routes to child nodes, defined as:
\begin{equation}
\label{BranchDistribution}
p_i(n_i=k|x) = \exp \left( \dfrac{w_k^T h^i_x + b_k}{\tau} \right) \bigg/ \sum_{j=1}^{K} \exp \left( \dfrac{w_j^T h^i_x + b_j}{\tau} \right)  \nonumber
\end{equation}
$h^i_x$ is the output feature of the router network at node $i$, usually the result of a fully connected layer. $(w_j,b_j)$ is the decision hyperplane for the $j$-th path. $\tau$ is used for smoothing the distribution; as $\tau \to \infty$, $p_i(n_i|x)$ converges to the uniform distribution and as $\tau \to 0$, it converges to the indicator function. It is usually initialized with a large number and annealed during training, which prevents the routers from being too confident with their decisions initially \cite{montillo2013entanglement}.
Using the joint distribution (\ref{Joint}), we define the information gain objective as the difference between the entropy of the class distribution at node $i$ and the expected entropy of the conditional class distributions after branching:
\begin{equation}
\label{InformationGain}
IG_{i} = \mathbb{H}\left[p_i(y)\right] -  \mathbb{E}_{p_i(n_i)} \left[ \mathbb{H}\left[p_i(y|n_i)\right]  \right]
\end{equation}
Here, entropy is defined as: $\mathbb{H}\left[p(x)\right] = -\sum_{x}p(x)\log p(x)$. An equivalent definition is the mutual information between the class label $y$ and the branch variable $n_i$: $IG_{i} = D_{KL}(p_i(y,n_i)||p_i(y)p_i(n_i))$, 
where $D_{KL}$ is the KL divergence. The derivative of the information gain with respect to network parameters is:
\begin{align}
\label{IG_derivative}
\nonumber
\diff{IG}{W} = &\sum_{n_i=1}^{K}\sum_{y=1}^{C}\mathbb{E}\left[p(y|x)\diff{p_i(n_i|x)}{W}\right]\log \mathbb{E}\left[p(y|x)p_i(n_i|x)\right]\\ \nonumber 
&-\sum_{n_i=1}^{K}\mathbb{E}\left[\diff{p_i(n_i|x)}{W}\right]\log \mathbb{E}\left[p_i(n_i|x)\right] \nonumber 
\end{align}
where the expectations are over $p_i(x)$. While this derivative has a biased estimator for minibatch updates, this does not cause any problems in SGD iterations and the information gain objectives converge fairly fast.
 
\subsection{Load Balancing}
The local information gain objective may converge to pure but unbalanced local minima in the sense that some leaves are assigned a high proportion of the classes where others get only a few. In such cases, the overloaded leaves may converge slowly, while others can quickly overfit with much fewer samples. To avoid such degenerate settings, we first decompose the information gain objective as in the following:
\begin{equation}
\label{IG_decomposed}
IG_{i} = \mathbb{H}\left[p_i(y)\right] + \mathbb{H}\left[p_i(n_i)\right] - \mathbb{H}\left[p_i(y, n_i)\right] \nonumber
\end{equation}
To increase the information gain, $\mathbb{H}\left[p_i(y)\right] + \mathbb{H}\left[p_i(n_i)\right]$ needs to be increased during training. $p_i(n_i)$ with high entropy enforces a balanced distribution, so we weight $\mathbb{H}\left[p_i(n_i)\right]$ with $\lambda_{balance} > 1$, such that the objective becomes:
\begin{equation}
\label{IG_decomposed}
IG_{i}^{balanced} = \mathbb{H}\left[p_i(y)\right] + \lambda_{balance}\mathbb{H}\left[p_i(n_i)\right] - \mathbb{H}\left[p_i(y, n_i)\right] \nonumber
\end{equation}
This new objective prefers optima with more balanced sample distributions, hence the coefficient $\lambda_{balance}$ acts a load balancing regularizer.

\subsection{Routing Errors}
A tree structured hierarchy is prone to routing errors: If a sample is routed to a path which is not an expert for its class, it may be misclassified. One approach could be improving the routing performance. Another approach would be to improve the generalization of the experts on samples which are misrouted to them and are not in the experts' respective data partition. We follow the second path. We route a sample $x$ in the split node $i$ into the $k$-th path, as long as $p_i(n_i=k|x) \geq \rho$, where $\rho$ is a threshold; starting with $0$ and slowly increased during the training. The experts gradually get focused on their partition, initially seeing data from other partitions and only focusing on their partitions later in the training. The upper bound for $\rho$ is $1/K$, where $K$ is the number of child nodes of node $i$. This ensures at least one path is always activated. During the test phase no thresholding is used; each sample is solely routed to the path $k$ with the highest $p(n_i=k|x)$.

\section{Experiments}
We tested our method on the MNIST \cite{lecun1998gradient} and newly introduced Fashion MNIST datasets \cite{xiao2017/online}. In our experiments, our approach was to take a CNN baseline, convert it into a CIGN, by introducing appropriate routing networks and decimating the number of total parameters in the classification pipeline. In each of our CIGNs, a root-leaf path network (expert) corresponds to a thinner version of the baseline CNN, excluding the additional router networks ($H$ transformations). For fair comparison, we only use the layer compositions found in the corresponding baseline networks in the CIGN models.

\subsection{MNIST}

\begin{table}[!t]
\renewcommand{\arraystretch}{1.2}
\caption{MNIST Test Results}
\label{MNIST_results}
\centering
\begin{tabular}{|c|c|c|c|c|}
\hline
\bfseries Model & \bfseries Max Ac. & \bfseries Min Ac. & \bfseries Avg Ac. & \bfseries \# of Params\\
\hline
LeNet Baseline & \%99.31 & \%99.22 & \%99.25 & 1256080\\
\hline
LeNet Baseline (*) & \%99.24 & \%99.16 & \%99.20 & 26695\\
\hline
CIGN, Ind. H & \%99.39 & \%99.29 & \%99.34 & 99856\\
\hline
CIGN, F fed to H & \%99.42 & \%99.28 & {\bf\%99.36} & 120366\\
\hline
DNDF, \cite{kontschieder2015deep} & - & - & \%99.3 & $\approx$785000\\
\hline
\end{tabular}

Each sample in the CIGNs visits a network equivalent to (*) plus routers.
\end{table}

\begin{figure}[!t]
\centering
\includegraphics[width=3.0in]{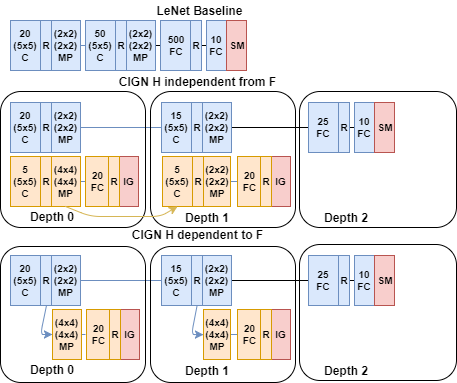}
\caption{The baseline LeNet and the CIGN architectures for MNIST. For CIGNs, a single root-leaf expert is shown. R stands for ReLU. For max pool operations the kernel-stride sizes are shown in the respective order. All convolutions have $(1 \times 1)$ stride.}
\label{fig:MnistArchitecture}
\end{figure}

We have first used the well known MNIST dataset. As the baseline, we use a version of the LeNet network, which uses rectifiers instead of sigmoid nonlinearities \cite{lecun1998gradient}. We used two CIGN versions, one using routers which are fed with the intermediate classification features ($F$ transformations) and one using an independent router system. We use binary trees with the depth of $2$ ($4$ root-leaf experts). We denote this structure with $[2,2]$. Figure \ref{fig:MnistArchitecture} shows the used network structures. Both for the baseline and CIGN, we have used a SGD optimizer with a momentum of $0.9$. The batch size is $125$ and the total number of epochs is $100$. The learning rate starts with $0.025$ and is halved at intervals of $15000$ iterations. We use $L2$-norm based weight decay regularizer and optimize its $\lambda_{baseline}$ with grid search, in the interval $[0.0,0.001]$ with a grid length of $5\times10^{-5}$. We similarly optimize $\lambda_F$ and $\lambda_H$; first for the CIGN with the independent $H$. Using the same grid search setting as the baseline, $\lambda_F$ is optimized first, then with the optimal $\lambda_F$ fixed, $\lambda_H$ is optimized. Optimal values are found to be $\lambda_{baseline}=9\times10^{-4}$, $\lambda_F=5\times10^{-5}$ and $\lambda_H=9\times10^{-4}$. We use these hyperparameters for the CIGN with the $F$ dependent routers, too, without grid search. $\lambda_{IG}$ is set as $1$. We experimented with $\lambda_{balance} \in \{1,2\}$ and determined $\lambda_{balance}=2$ leads to a more balanced class distribution. $\tau$ starts with $25$ and is decayed with $0.9999$ once in $2$ iterations up to a minimum of $\tau=1$. Finally we set $\rho=0$, allowing all samples to route into each path for $25$ epochs and then set it to $\rho=0.4$, letting only a very small (about \%2) percentage of samples to route to both paths at indecisive split nodes. Other than the grid search for $\lambda_{F},\lambda_{H}$ and for $\lambda_{balance}$ in $\{1,2\}$, no optimization has been applied to other hyperparameters. No augmentation has been used on the training data. Finally, we run the baseline and both CIGN models with their determined settings on the test set, 5 times for each. The results of the experiments are shown in Table \ref{MNIST_results}, with best accuracies in bold. We see that the CIGN with routers connected to the classification pipeline gives the best results. Our results also compare favorably with DNDF \cite{kontschieder2015deep}, which uses the same LeNet architecture on unaugmented MNIST data, by replacing the softmax with their decision forest classifier. We also give the results of the thin version of the LeNet architecture, on which we have applied the same grid search protocol. This is the version which we use as the root-leaf network in our CIGN model. It is clear that our better results are not only due to avoiding overfitting with lesser number of parameters, since the reduced version cannot surpass the original baseline in terms of accuracy.

\subsection{Fashion MNIST}
\begin{figure}[!t]
\centering
\includegraphics[width=3.5in]{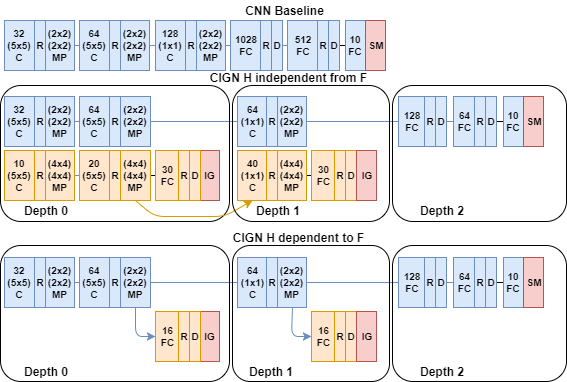}
\caption{The baseline and CIGN models we used in Fashion MNIST experiments. Same notations applies as Figure \ref{fig:MnistArchitecture}. D stands for Dropout layer.}
\label{fig:FashionMnistArchitecture}
\end{figure}

The recently introduced Fashion MNIST contains similar sized data as MNIST ($28\times28$ grayscale images with 60000 training and 10000 test samples) \cite{xiao2017/online}. It contains 10 different types of clothing: T-shirt, Trouser, Pullover, Dress, Coat, Sandal, Shirt, Sneaker, Bag and Ankle Boot. It is more challenging compared to MNIST. Therefore, we use a larger CNN as our baseline, which contains 3 convolutional and 2 fully connected layers (Figure \ref{fig:FashionMnistArchitecture}). We again use $[2,2]$ CIGNs, one with independent routers and another with routers connected to the intermediate classification features. No augmentation has been used. As the optimizer, we again use SGD with a momentum of 0.9. The training for all models continue for 100 epochs. The batch size is 125 and the learning rate starts as $0.01$. It is halved at $15000$. and $30000$. iterations and multiplied with $0.1$ at $40000$. iteration. For regularization, we use Dropout and insert it after each fully connected layer. Both for the baseline and classification pipelines of the CIGNs, we conduct a grid search for the dropout probability in the interval $[0,0.5]$ with the grid length of $0.05$. We determine the optimal values for the baseline and for the CIGNs with dependent and independent routers, respectively, as $0.35$, $0.15$ and $0.2$. We also insert a dropout layer before each information gain objective in every split node, for CIGNs. Without grid search, we set its dropping probability as $0.35$. For Fashion MNIST, a larger $\lambda_{balance}$ is needed to find a balanced sample distribution. We search on $\lambda_{balance} \in \{2,3,4,5\}$ and determine that $\lambda_{balance}=5$ achieves a reasonably balanced and pure partitioning of the data. We set $\lambda_{IG}=1$ and use the same annealing processes both for $\tau$ and $\rho$ as in the MNIST experiments. The results are given in Table \ref{Fashion_MNIST_results}, which are the average of 6 runs. We again give the reduced root-leaf network's accuracy, which is inferior compared to the original baseline and our CIGN structures.

\begin{figure}[!t]
\centering
\includegraphics[width=3.5in]{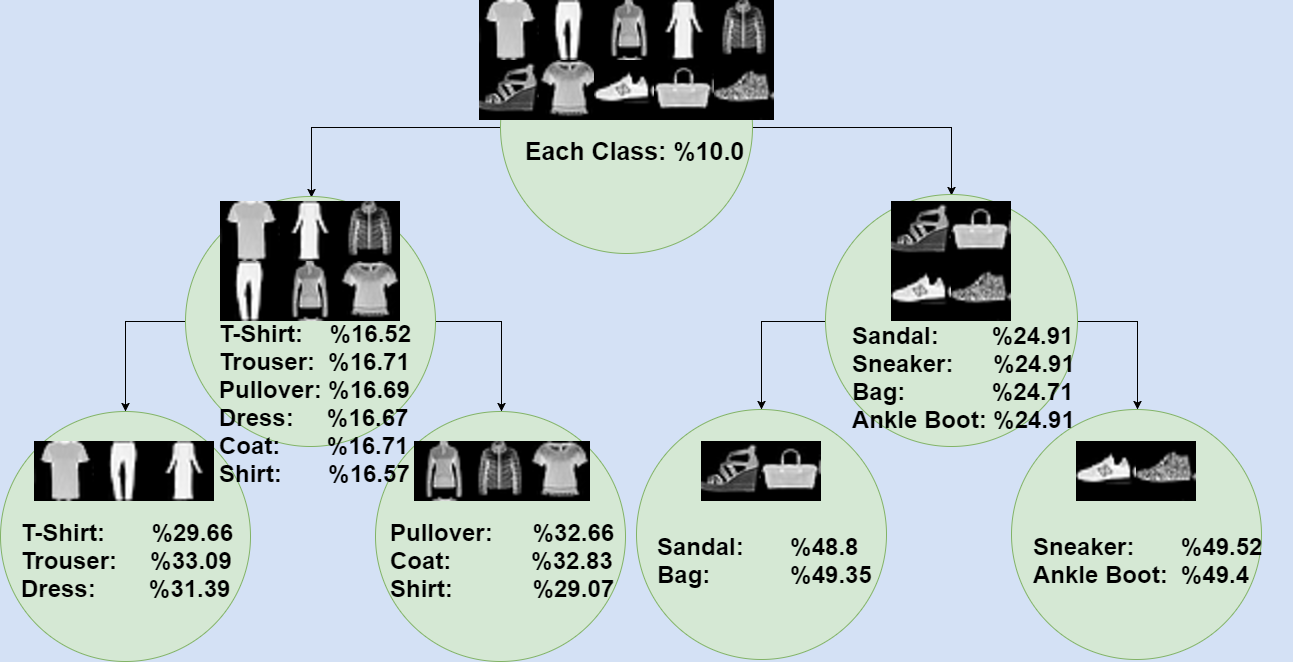}
\caption{A tree shaped histogram, showing how the Fashion MNIST test data is partitioned by a $[2,2]$ CIGN; as taken from the result of an actual experiment run. We omit classes with very low frequencies from the nodes for the sake of clarity. Note that Fashion MNIST is a strictly balanced dataset with each class having the same training and test sample count as other classes.}
\label{fig:TreeHistogram}
\end{figure}

It is also interesting to examine how the data is partitioned in a $[2,2]$ tree. Fashion MNIST has an instrinsic modality; there are three footwear, six clothing classes and the Bag class, which do not strictly belong to these two groups. Figure \ref{fig:TreeHistogram} shows the routing decisions of CIGN on the Fashion MNIST test set. We observe that the first split node separates shoes and bag from clothing. Similarly, the child nodes group similar classes together (Trouser and Dress, Sneaker and Ankle Boot) and train classifiers that specialize in classifying between these similar classes.

\begin{table}[!t]
\renewcommand{\arraystretch}{1.2}
\caption{Fashion-MNIST Test Results}
\label{Fashion_MNIST_results}
\centering
\begin{tabular}{|c|c|c|c|c|}
\hline
\bfseries Model & \bfseries Max Ac. & \bfseries Min Ac. & \bfseries Avg Ac. & \bfseries \# of Params\\
\hline
CNN Baseline & \%92.61 & \%92.00 & \%92.27 & 2688522\\
\hline
CNN Baseline (*)& \%92.22 & \%91.70 & \%91.96 & 196362\\
\hline
CIGN, Ind. H & \%92.59 & \%92.19 & \%92.32 & 643016\\
\hline
CIGN, F fed to H & \%92.52 & \%92.26 & \textbf{\%92.36} & 713736\\
\hline
\end{tabular}

Each sample in the CIGNs visits a network equivalent to (*) plus routers.
\end{table}

\section{Conclusion}
In this paper, we have presented Conditional Information Gain Networks (CIGN), which allow end-to-end training of tree structured, conditional deep neural networks. The conditioning is due to the decision mechanisms that are built on differentiable, local information gain objective functions. This structure allows both the training of expert root-leaf networks specialized on a subset of classes, with significantly lower number of parameters compared to a baseline CNN and conditional computation during inference and training. Our tests with the MNIST and Fashion MNIST datasets give indicators about the effectiveness of our method.

As future fork, we plan to adapt the method to larger CNN models like ResNets \cite{He_2016_CVPR} and to work on more complex datasets. Especially datasets with multiple labels per sample would be an interesting choice, since there can be numerous ways to exploit the rich label structure with our information gain based decision mechanism.

\section*{Acknowledgment}
This work is supported by the Turkish Ministry
of Development under the TAM Project, number 2007K120610.




%


\bibliographystyle{IEEEtran}
\bibliography{IEEEabrv,refs}

\end{document}